\documentclass[10pt,twocolumn,letterpaper]{article}

\usepackage{cvpr}
\usepackage{times}
\usepackage{epsfig}
\usepackage{graphicx}
\usepackage{amsmath}
\usepackage{amssymb}
\usepackage{array}
\usepackage{tabu}
\usepackage{multirow}
\usepackage{makecell}

\usepackage{placeins}

\usepackage[pagebackref=true,breaklinks=true,letterpaper=true,colorlinks,bookmarks=false]{hyperref}

\cvprfinalcopy 

\def\cvprPaperID{11} 

\ifcvprfinal\fi

\makeatletter
\newcommand{\thickhline}{%
    \noalign {\ifnum 0=`}\fi \hrule height 1pt
    \futurelet \reserved@a \@xhline
}
\newcolumntype{"}{@{\hskip\tabcolsep\vrule width 1pt\hskip\tabcolsep}}
\makeatother

\begin{document}

\title{Object Level Deep Feature Pooling for Compact Image Representation}

\author{Konda Reddy Mopuri and R. Venkatesh Babu\\
Video Analytics Lab, SERC,\\
Indian Institute of Science, Bangalore, India.\\
{\tt\small sercmkreddy@ssl.serc.iisc.in, venky@serc.iisc.ernet.in}
 }

\maketitle

\begin{abstract}
Convolutional Neural Network (CNN) features have been successfully employed in recent
   works as an image descriptor for various vision tasks. But the inability of the deep CNN features to exhibit invariance
   to geometric transformations and object compositions poses a great challenge for image search. In this work, we demonstrate
   the effectiveness of the objectness prior over the deep CNN features of image regions for obtaining an invariant image representation.
   The proposed approach represents the image as a vector of pooled CNN features describing the underlying objects. This representation 
   provides robustness to spatial layout of the objects in the scene and achieves invariance to general geometric 
   transformations, such as translation, rotation and scaling. The proposed approach also leads to a compact representation of 
   the scene, making each image occupy a smaller memory footprint. Experiments show that the proposed representation achieves 
   state of the art retrieval results on a set of challenging benchmark image datasets, while maintaining a compact representation.
\end{abstract}

\section{Introduction}
Semantic image search or content based image retrieval is one of the well studied
problems in computer vision. Variations in appearance, scale and orientation changes and change
in the view point pose a major challenge for image representation. In addition, the huge  volume of 
image data over the Internet adds to the constraint that the representation should also be compact for
efficient retrieval.
%
\begin{figure}[ht]
 \includegraphics[width=0.49\textwidth]{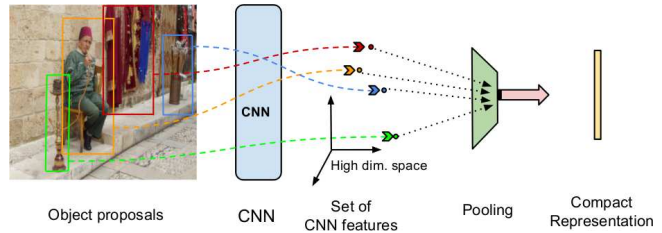} %
 \caption{\textbf{The overview of the proposed system.} First stage obtains 
 object proposals from the input image. Second stage extracts deep features from these regions.
 Finally, the last stage pools these features in order to obtain  a compact representation.}
 \label{fig:sys}
\end{figure}

Typically, an image is represented as a set of local features for various computer vision
applications. Bag of Words (BOW) model~\cite{sivic2003video} is a well studied
approach for content based image retrieval (CBIR). The robustness
of the SIFT like local features~\cite{sift}~\cite{surf}~\cite{mser} to various geometric transformations and applicability
of different distance measures for similarity computation have led to a widespread adoption of this framework.

The BOW model is inspired from the familiar Bag of Words approach for document retrieval.
In this model, an image is represented as a histogram over a set of learned codewords after quantizing
each of the image features to the nearest codeword. However, computing
the BOW representation for an image is tedious and the computation time grows linearly with the size of the codebook.
Although BOW model provides good accuracy for image search, it is not scalable
for large image databases, as it is intensive both in memory and computations.

In order to have a more informative image representation, Peronnin \textit{et al.} proposed 
a generative model based approach called Fisher vector \cite{fisher1}. They estimate a
parametric probability distribution over the feature space from a huge representative set of
features, generally a gaussian mixture model (GMM). The features extracted from an image are
assumed to be sampled independently from this probability distribution.
Each feature in the sample is represented by the gradient of the probability distribution at that
feature with respect to its parameters. Gradients corresponding to all the features with respect
to a particular parameter are summed. The final image representation is the concatenation of the accumulated gradients. They
achieve a fixed length vector from a varying set of features that can be used in various
discriminative learning activities. This approach considers the $1^{st}$ and $2^{nd}$ order
statistics of the local image features as well, whereas the BOW captures only the $0^{th}$
order statistics, which is just their count.


J\'{e}gou \textit{et al.} proposed Vector of Locally Aggregated Descriptors (VLAD)~\cite{vlad2012pami} as a 
non-probabilistic equivalent to Fisher vectors. VLAD is a simplified and special case of Fisher vector, in which
the residuals belonging to each of the codewords are accumulated. The concatenated residuals represent the image, 
using which the search is carried out via simple distance measures like $l_2$ and $l_1$ . A number of 
improvements have been proposed to make VLAD a better representation by considering vocabulary adaptation and 
intra-normalization~\cite{allaboutvlad}, residual normalization and local coordinate 
system~\cite{revisitingvlad}, geometry information ~\cite{gvlad} and multiple 
vocabularies~\cite{multivocvlad}.

All these approaches are built on top of the invariant SIFT-like local features. On the other hand,
features 
like GIST~\cite{gist} are proposed towards an image representation via a global image feature. 
Douze \textit{et al.}~\cite{gisteval} attempted to identify the cases in which a
global description can reasonably be used in the context of object recognition and copy detection.

Recently, the deep features extracted from the Convolutional Neural networks (CNN) 
have been observed \cite {decaf,midlevelcnn} to show a better performance 
over the the state-of-the-art for important vision applications such as object detection, 
recognition and image classification \cite{rcnncvpr2014}. This has inspired 
many researchers to explore deep CNNs in %
order to solve a variety of problems~\cite{nc14, rcnncvpr2014, 	mopcnneccv14}. 

For obtaining an image representation using CNNs, the mean-subtracted image is forward propagated
through a set of convolution layers. Each of these layers contains filters that convolve the outputs 
from previous layer followed by max-pooling of the resulting feature maps within a local neighborhood.
After a series of filtering and sub-sampling layers, another series of fully connected layers process 
the feature maps and leads to a fixed size representation in the end. Similar to the multi layer 
perceptron (MLP), the output at each of the hidden units is passed through a non-linear activation
function to induce the nonlinearity into the model. 

Because of the local convolution operations, the representation preserves the 
spatial information to some extent. For example, Zeiler \textit{et al.} \cite{visualizingcnn} 
have shown that the activations after the max-pooling of the fifth layer can be reconstructed
to a representation that looks similar to the input image. This makes the representation to be sensitive to
the arrangement of the objects in the scene. Though max-pooling within each feature map contributes
towards the invariance to small-scale deformations~\cite{Leeicml}, transformations that are more
global are not well-handled due to the preserved spatial information in the activations.

Intuitively, the fully connected layers, because of their complexity, are supposed to provide the highest
level of visual abstraction. For the same reason, almost all the works
represent the image content with the activations obtained from the fully connected layers.
There is no concrete reasoning provided to comment on the invariance properties of the 
representations out of the fully connected layers. However, Gong \textit{et al.}\cite{mopcnneccv14} have empirically 
shown that the final CNN representation is affected by the transformations such as scaling, rotation and translation.
In their experiments, they report that this inability of the activations has been translated into direct a loss of 
classification accuracy and emphasize that the activations preserve the spatial information.
As another evidence, Babenko \textit{et al.}~\cite{nc14} show that the retrieval 
performance of the global CNN activations (called `neural codes') increases
considerably, when the rotation in the reference (database) images is nullified manually.

\begin{figure}[ht]
\centering
 \includegraphics[width=0.4\textwidth]{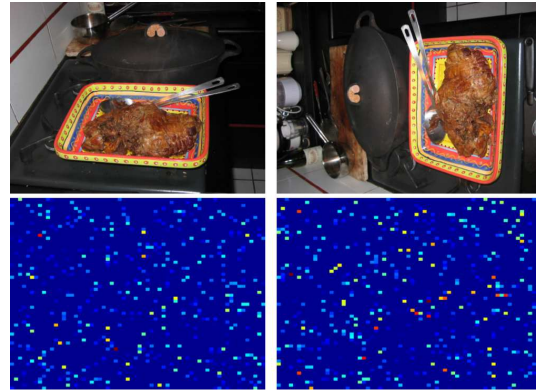} 
 \caption{Sensitivity of the CNN features to global transformation. Top row: two images from the
 Holidays dataset, with the same objects  but re-arranged.
 Bottom row: Corresponding $l_2$ normalized $4096D$ CNN features shown on $64 \times 64$ grid
 (Best viewed when zoomed-in).}
 \label{fig:sensitivity}
\end{figure}

We have considered two images of the same scene from
the Holidays database. We have conducted an experiment
to demonstrate the sensitivity of the CNN features to global
transformation that results in a different arrangement of the
underlying objects. We extracted the $4096$ dimensional
CNN feature for both the images after the first fully
connected layer (fc6) in the Alexnet, as described in~\cite{deepcnnnips2012}.
Figure \ref{fig:sensitivity} shows the two images and the corresponding deep features reshaped to 
a $64 \times 64$ matrix. We can clearly observe that the corresponding activations to be very 
different in both the representations, leading to an inferior similarity between the two images.


Gong \textit{et al.}\cite{mopcnneccv14} proposed a simple framework called Multi-scale Orderless Pooling (MOP-CNN)
towards an orderless representation via pooling the activations from a set of patches extracted at different scales.
They resized the image to a fixed size and considered patches at three fixed scales in an exhaustive manner to extract CNN
features from each of the patches. The resulting CNN features are VLAD pooled (aggregated) in
order to achieve an image representation.

Motivated by this approach, we construct an image representation on top of the CNN activations 
without binding any spatial information. The proposed approach, constructs a novel object based
invariant representation on top of the CNN descriptions
extracted from an image. We describe the objects present in an image through the deep features and 
pool them into a representation, as to keep the effective visual content of the image.

Generating object proposals is one of the effective solutions in the recent times to achieve 
computational efficiency for object detection. Our proposed system employs the selective search object
proposals scheme developed by 
Uijlings \textit{et al.} \cite{selectivesearchijcv2013} to extract regions from the 
given image. Selective search relies on a bottom-up grouping approach for segmentation
that results in a hierarchical grouping. Hence, it naturally enables to generate
object proposals at all scales.

After obtaining the object level deep features, constructing an image representation on top of them is another challenge.
In order to have a holistic image representation that summarizes the entire visual content, we
propose to keep the largest of the activations at each of the output units in the CNN. That means the representation is
the max pooled output vector of all the object level CNN activations extracted from an image.
We have experimentally observed that the proposed representation consistently
outperforms other representations based on deep features (refer section \ref{sec:ER}).

Figure \ref{fig:sys} shows the overview of the proposed system. The reminder of the paper is organized as follows:
In section \ref{sec:PA}, we discuss each of 
the modules of the proposed retrieval system in detail. In section \ref{sec:ER}, we present the
experimental setup using which we evaluate the invariance of the proposed image representation
derived from the object level CNN activations and discuss the results. Section \ref{sec:conclu}
summarizes our approach with a conclusion and possible expansions.

\section{Proposed Approach}
\label{sec:PA}
In this paper, we propose to use objectness prior on the Convolutional Neural Network
(CNN) activations obtained from the image in order to represent the image compactly. 
Our proposed system as shown in Figure \ref{fig:sys}, is essentially a cascade of three modules. 
In this section we discuss these three modules in detail.

\subsection{Region Proposals}
The first stage of the proposed system is to obtain a set of object regions from the image. 
Recent research offers a variety of approaches~
\cite{selectivesearchijcv2013}~\cite{bing}~\cite{edgeboxes}
for generating a set of class independent region proposals on an image. The motivation for 
this module comes from the fact that tasks such as detection and
recognition rely on the localization of the objects present in the image. Thus the approach
is an object based representation learning built on the deep features.

 The conventional approach to object detection
has been an exhaustive search over the image region with a sliding window. 
The recent alternative framework of 
objectness proposals aims to propose a set of object bounding boxes with an 
objective of reducing the complexity in terms of the image locations needed to be further
analyzed. Another attractive feature of the object proposals framework is that, the 
generated proposals are agnostic to the type of the objects being detected.

In the proposed system, we adapt the region proposals approach by Uijlings \textit{et al.} 
\cite{selectivesearchijcv2013}. This method combines the strength 
of both an exhaustive search and segmentation. This method relies on a bottom-up hierarchical segmentation
approach, enabling itself naturally to generate regions at different scales. The object
regions obtained are forwarded through a CNN for an efficient description.

Since the objects in the scene are crucial components to understand the underlying scene, 
we use object based representation for describing the scene. Further, in semantically similar scenes, these
objects need not be present in the same spatial locations, making the object level description a necessity. 
Unlike MOP-CNN~\cite{mopcnneccv14} which preserves the geometry information loosely
because of the concatenation, the proposed approach does not encode
any geometric information into the final representation. 
Wei \textit{et al.}~\cite{HCP} proposed Hypotheses-CNN-Pooling (HCP),  an object based multi-label image classification approach that 
aggregates the labels obtained at object level classification. They obtain objects present in the image as different hypotheses and
classify them separately using a fine-tuned CNN. The final set of labels are obtained as the result of max-pooling the outputs 
corresponding to the hypotheses, emphasizing the importance of the objects in the scene.


\subsection{Feature Extraction: CNN features }

The objective of the CNN in the proposed system is to achieve a high level semantic description of 
the input image regions. For over more than a decade, the vision community has witnessed the dominance enjoyed
by a variety of hand-crafted SIFT like\cite{sift,surf,hog} features for applications like object recognition and retrieval. 
But a number of recent works~\cite{decaf,rcnncvpr2014,mopcnneccv14,midlevelcnn} have claimed the
supremacy of the features extracted from the Convolutional Neural Networks (CNN) and Deep learning over the 
conventional SIFT-like ones.

CNNs are biologically inspired frameworks to emulate cortex-like mechanisms and the architectural depth of the brain
for efficient scene understanding. The seminal works by LeCun \textit{et al.}~\cite{lenet98} has inspired many researchers
to explore this framework for many vision tasks. A similar architecture proposed by Serre \textit{et al.}~\cite{serre} 
also tries to follow the organization of the visual cortex and builds an increasingly complex and invariant representation.
Their architecture alternates between template matching and max-pooling operations and is capable of learning from 
only a few set of training samples. 

Compared to the standard feedforward 
networks with similar number of hidden layers, CNNs have fewer connections and parameters, making it easier to learn. 
In the proposed approach, we describe each of the extracted image regions with a fixed
size feature vector obtained from a high capacity CNN. Our feature extraction is similar to that in \cite{rcnncvpr2014}.
 We have used the Caffe \cite{caffe} implementation of the CNN described by Krizhevsky \textit{et al.} \cite{deepcnnnips2012}.
We pass the image region through five convolution layers, each of which is followed by a sub-sampling (or pooling) and 
a rectification. After all the convolution layers, the activations are passed through a couple of fully connected 
layers in order to extract a 4096-dimensional feature vector. More about the architecture details
of the network can be found in \cite{caffe}~and~\cite{deepcnnnips2012}.

\subsection{Feature Pooling}
\label{ss:FP}
At this stage, each image is essentially a set of object level CNN features. The extracted features from the CNN
are max-pooled in order to obtain a compact representation for the input image.

Apart from the obvious aggregation to deliver a compact representation, the main motivation for the
max-pooling stems form the aim to construct a representation that sums the visual content of the
image. It is also observed by Girshick \textit{et al.}~\cite{rcnncvpr2014} that, at the intermediate layers of the CNN
(like pool $5$ layer in~\cite{rcnncvpr2014}), each of the units responds with a high activation to a specific type of visual input. 
The specific image patches causing these high activations are observed to belong to a particular object class. It 
can be thought of as, each of the receptive fields in the 
network is probing the image region for the content of a specific semantic class. Thus, keeping 
the component-wise maximum of all the CNN activations belonging to the object proposals sums up the whole visual
content of the image.
%
%

\subsection{Query time Processing}
\label{subsec:QP}
During the query time, we apply selective search on the given query image to extract object proposals. These proposals 
are propagated forward through the CNN in order 
to describe them with a fixed length $(4096D)$ representation. At this point, we rank the proposals
using their objectness scores. We use a Jaccard intersection-over-union measure to reject
the lower scoring proposals and retain the higher scoring ones.


Finally, we employ max-pooling to aggregate all the CNN activations in order to 
represent the visual content of the input image. We compare the pooled vector representation of the query with that of the 
database images to retrieve the similar images. Simple distance measures such as $l_2$ and Hamming distance (for the binarized
representations of the pooled output) are found suitable to compare the images.
\section{Experiments and Results}
\label{sec:ER}
\begin{figure}
\centering
\includegraphics[width=3.2in]{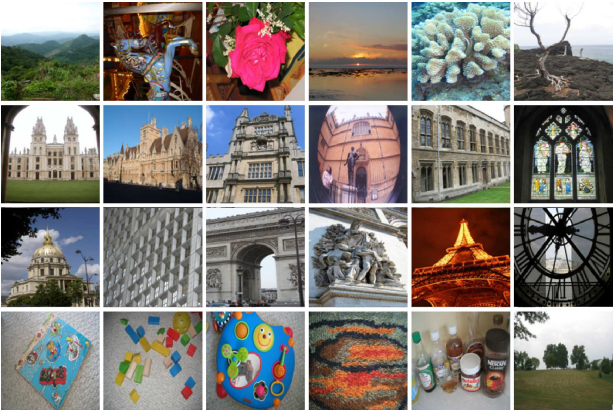} 
\caption{Sample database images. First row shows images from the \emph{Holidays}, second from the \emph{Oxford5K}, 
third from the \emph{Paris6K} and the fourth from the \emph{UKB} dataset.}
\label{fig:SampleImags}
\end{figure}

In this section, we demonstrate the effectiveness of having an objectness prior on the Convolutional
Neural Network (CNN) activations obtained from the object patches of an image in order to achieve an 
invariant image representation. We evaluate these invariant representations via a series of
retrieval experiments performed on a set of benchmark datasets. Note that our primary aim is 
to have a compact representation for the images at both query and database ends over which we 
build an image retrieval system. We evaluate our approach over the following publicly available
benchmark image databases. Sample images from the databases are shown in Fig. \ref{fig:SampleImags}.
\subsection{Datasets}
{INRIA Holidays\cite{jegou2008hamming}}:
This dataset contains $1491$ photographs captured at different places that fall into 
$500$ different classes. One query per class is considered for evaluating the proposed system through Mean Average Precision
(mAP).

{Oxford5K\cite{philbin2007object}}:
This database is a collection of $5062$ images of $11$ landmark buildings in the oxford 
campus that are downloaded from Flickr. 55 hold-out queries, uniformly distributed over the $11$
landmarks evaluate the retrieval performance through Mean Average Precision (mAP).

{Paris6K\cite{paris}}:
The \emph{Paris} dataset consists of around $6400$ high resolution images collected from Flickr 
by searching for $11$ particular landmarks in the city of Paris. Similar to the \emph{Oxford} 
database, 55 queries distributed over these $11$ classes. Mean Average Precision (mAP)
is considered as the metric for evaluation.

{UKB\cite{ukb}}:
\emph{UKB} dataset contains $10200$ images of $2550$ different indoor objects. This can be considered as having 
$2550$ classes each populated with four images. We have considered One query image per class and
four times the precision at rank $4$ $($4$ \times precision@4)$ as the evaluation metric 
for this database.

\subsection{Compact Representation and Retrieval}
The proposed approach extracts the possible object patches as suggested by the 
selective search \cite{selectivesearchijcv2013} approach. Because of the objectness prior associated with
each of these patches, our approach is more accurate and 
reliable in aggregating the effective visual content of the image.
From each image, we extract on an average $2000$ object proposals. These patches are 
described by the $4096$ dimensional features obtained using a trained CNN. Through max-pooling, 
we retain the component-wise maxima of all these features. The resulting
representation is still a $4096$ dimensional vector that sums up the visual information present 
in the image. We use the Principal Component Analysis (PCA) to reduce the dimensionality of 
the representation. The retrieval is conducted using these lower dimensional representations
on all the databases and the results are presented in Tables~\ref{tab:mapResHD}, \ref{tab:mapResOX},
\ref{tab:mapResPR} and \ref{tab:mapResUKB}. We compare the performance of the proposed approach with 
the existing methods for various sizes of the compact codes. 

\begin{center}
\begin{figure*}

\includegraphics[width=\textwidth,height=3.5in]{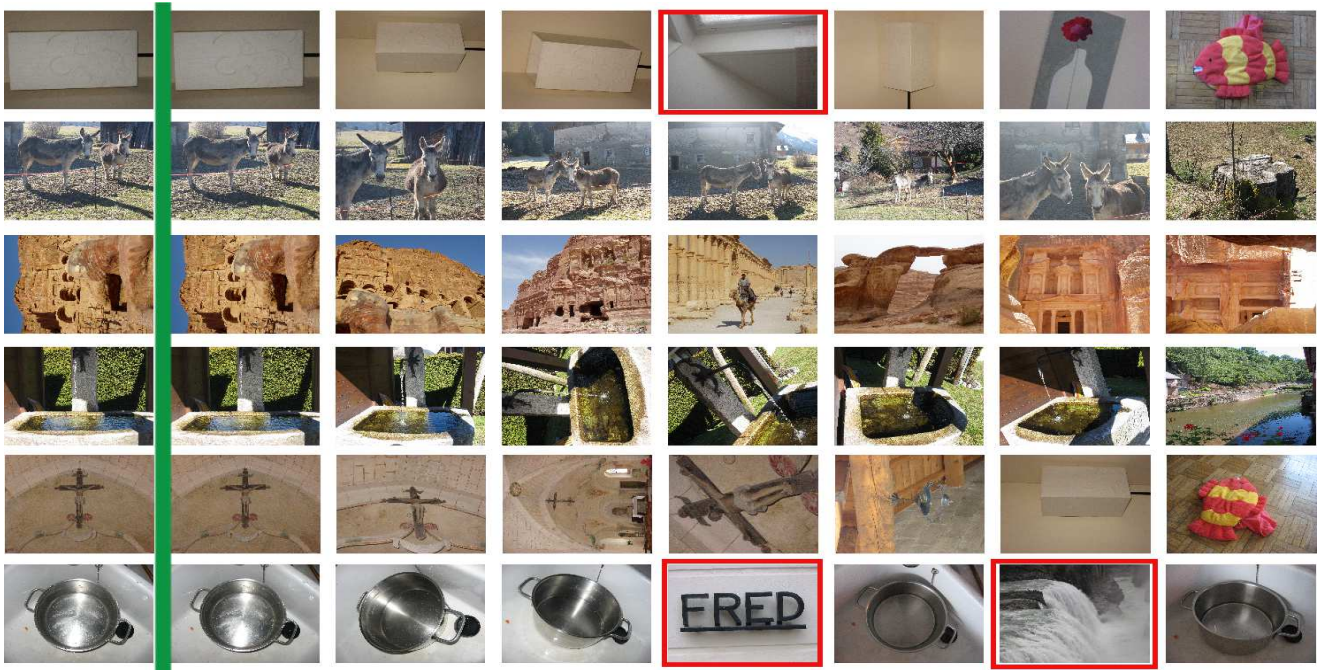} 

\caption{\textbf{Retrieval results on Holidays.} First image in each row (separated by green line) is the query image and the 
subsequent images are the corresponding ranked images. The red boxes show images that are irrelevant and ranked higher than 
the relevant images. Note that irrelevant images ranked after the relevant images are not shown in red boxes. Note that queries 
$1, 3,4$ and $5$ are picked to demonstrate the invariance to rotation achieved by the proposed approach. Note that, the query image is 
removed from the ranked list while computing the mAP.}
\label{fig:retresults}
\end{figure*}
\end{center}



\begin{table*}[t]
\caption{Retrieval results on the \emph{Holidays} dataset. Best performances in each column are shown in bold.
(\textbf{$\nabla$} \textit{indicates result obtained with manual geometric alignment and 
retraining the CNN with similar database}.)}
\centering
\begin{tabular}{lcccccccccc}
\hline\\[-7pt]
&        \multicolumn{9}{c}{Dimension} \\ \cline{2-11}\\[-5pt]
{ \textsc{Method}} 					 &32&64&128&256&512&1024& 2048  & 4096 &8064&$\geq10K$ \\ \hline \\
VLAD \cite{vlad2012pami}    &48.4&52.3&55.7&-&59.8&-&62.1&55.6&&\\
Fisher Vector\cite{fisher1}	&48.6&52&56.5&-&61&-&62.6&59.5&&\\
VLAD +adapt+ innorm \cite{allaboutvlad}&-&-&62.5&-&-&-&-&-&-&64.6\\
Fisher+color \cite{cvpr12} 	&-&-&-&-&-&-&-&77.4&	\\ 
Multivoc-VLAD \cite{multivocvlad}&-&-&61.4&-&-&-&-&-\\
Triangulation Embedding~\cite{triangulation}&-&-&61.7&-&-&72.0&-&-&77.1 \\

Sparse-coded Features~\cite{sparsecoded}&-&-&0.727&-&-&-&-&-&-&76.7\\
Neural Codes \cite{nc14}	&68.3&72.9&{78.9$^\nabla$}&74.9&74.9&-&-&\textit{79.3$^\nabla$} \\ 
MOP-CNN \cite{mopcnneccv14}	&-&-&-&-&-&-&80.2&78.9& 	\\ 
gVLAD \cite{gvlad}&-&-&77.9&-&-&-&-&81.2&\\ \hline\\[-5pt]
Proposed    	    &\textbf{73.96}&\textbf{80.67}&\textbf{85.09}&\textbf{87.77}&\textbf{88.46}&\textbf{86.58}&\textbf{85.94}&\textbf{85.94}& \\ \hline

\end{tabular}
\label{tab:mapResHD}
\end{table*}



\begin{table*}[t]
\caption{Retrieval results on the \emph{Oxford5K} dataset. Best performances in each column are shown in bold. 
(\textbf{*} indicates \textit{result obtained with retraining the CNN with similar database}.)}
\centering
\begin{tabular}{lccccccccc}
\hline\\[-7pt]
&        \multicolumn{9}{c}{Dimension} \\ \cline{2-10}\\[-5pt]
METHOD	&32&64&128&256&512&1024& 2048  & 4096 &8064 \\ \hline \\

VLAD \cite{vlad2012pami}    &-&-&28.7&-&-&-&-&-		\\
Fisher Vector\cite{fisher1}	&-&-&30.1&-&-&-&-&-\\
Neural Codes \cite{nc14}&39.0&42.1&43.3&43.5&43.5&-&-&\textit{54.5$^\textbf{*}$}\\ 
VLAD +adapt+ innorm \cite{allaboutvlad}&-&-&44.8&-&-&-&-&55.5\\
gVLAD \cite{gvlad}&-&-&\textbf{60}&-&-&-&-&\textbf{62.6}\\ 
Triangulation Embedding~\cite{triangulation}&-&-&43.3&-&-&56.0&57.1&62.4&\textbf{67.6}\\ \hline \\[-5pt]
Proposed    	    &\textbf{40.1}&\textbf{48.02} &56.24 &\textbf{59.78}&\textbf{60.71}&\textbf{59.42}&\textbf{58.92}&58.20    	\\ \hline

\end{tabular}
\label{tab:mapResOX}
\end{table*}



\begin{table*}[t]
\caption{Retrieval results on the \emph{Paris6K} dataset. Best performances in each column are shown in bold. (\textbf{*} \textit{ 
indicates result obtained with retraining the CNN with similar database}.)}

\centering
\begin{tabular}{lcccccc cc}
\hline\\[-7pt]
&        \multicolumn{8}{c}{Dimension} \\ \cline{2-9}\\[-5pt]
METHOD      				&32&64&128&256&512&1024& 2048  & 4096  \\ \hline \\
VLAD \cite{vlad2012pami}    &-&-&28.7&-&-&-&-&-		\\
Fisher Vector\cite{fisher1}	&-&-&30.1&-&-&-&-&-\\
Neural Codes~\cite{nc14}&39.0&42.1&43.3&43.5&43.5&-&-&54.5 \textbf{*}\\ 
gVLAD \cite{gvlad}&--&-&59.2&-&-&-&-&63.1\\ \hline \\[-5pt]
Proposed    	    &\textbf{65.38}&\textbf{71.47}&\textbf{70.39}&\textbf{68.43}&\textbf{66.23}&\textbf{64.11}&\textbf{62.84}&\textbf{63.20}    \\ \hline

\end{tabular}
\label{tab:mapResPR}
\end{table*}


\begin{table*}[t]
\caption{Retrieval results on the UKB dataset. Best performances in each column are shown in bold. (\textbf{*} \textit{indicates result obtained with  
retraining the CNN with similar database}.)}
%
\centering
\begin{tabular}{lcccccccccc}
\hline\\[-7pt]
&        \multicolumn{10}{c}{Dimension} \\ \cline{2-11}\\[-5pt]
METHOD      				&32&64&128&256&512&1024& 2048  & 4096&8064& $\geq 10K$  \\ \hline \\
VLAD~\cite{vlad2012pami}    &-&-&3.35&-&-&-&-&-&-&-		\\
Fisher Vector~\cite{fisher1}	&-&-&3.33&-&-&-&-&-&-&-\\
Triangulation Embedding~\cite{triangulation} &-&-&3.4&3.45&3.49&3.51&-&-&	3.53 	&-\\ 
Neural Codes~\cite{nc14}&3.3\textbf{*}&3.53\textbf{*}&3.55\textbf{*}&3.56\textbf{*}&3.56\textbf{*}&-&-&3.56 \textbf{*}&-&-\\ 
Sparse-Coded Features~\cite{sparsecoded}&\textbf{3.52}&\textbf{3.62}&3.67&-&-&-&-&-&-&3.76 \\ \hline \\[-5pt]
Proposed    	    &3.4&3.61&\textbf{3.71}&\textbf{3.77}&\textbf{3.81}&\textbf{3.84}&\textbf{3.84}&\textbf{3.84} &   -&-	\\ \hline
\end{tabular}
\label{tab:mapResUKB}
\end{table*}
%
%
\subsection{Binarization}
A great deal of compression can be achieved via binary coding. We adopt the recent work, Iterative Quantization (ITQ) by 
Gong \textit{et al.} \cite{itq} to obtain binary representation
for our pooled representation while preserving the similarity. The method proposes a simple iterative optimization strategy 
to rotate a set of 
mean centered data so that the quantization error is minimized to assign each point to vertices of a unit binary cube, in the 
projected space of arbitrary dimension.

\begin{figure}
\centering
\includegraphics[width=0.51\textwidth]{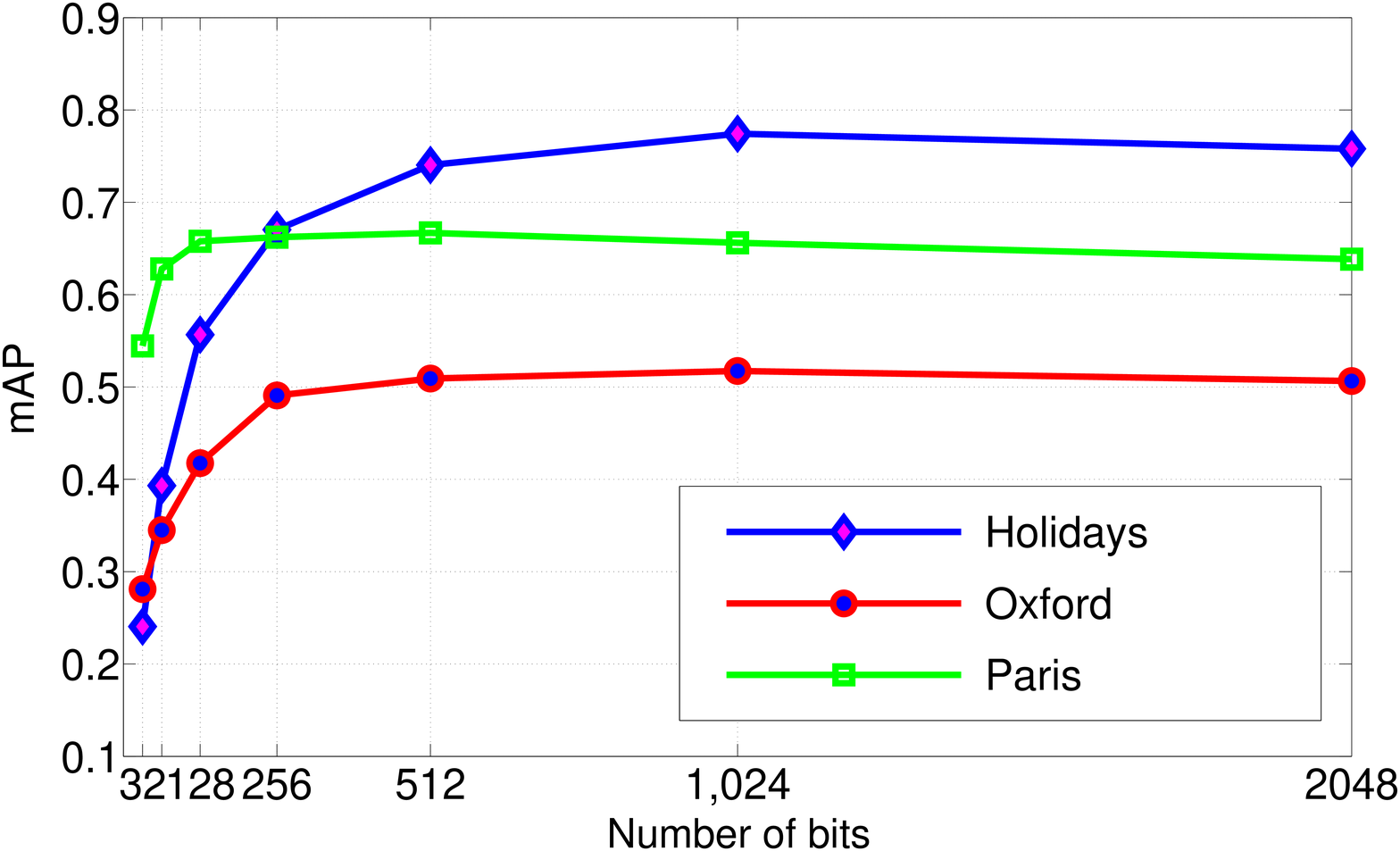} 
\caption{Retrieval performance on various datasets for different number of bits in the binary representation obtained via ITQ.}
\label{fig:mapvsITQ}
\end{figure}

The results presented in Figure \ref{fig:mapvsITQ} show that the binary representation retains the
retrieval performance in spite of the very compact representation.

\subsection{Number of Proposals}
In this subsection, we investigate the retrieval performance as a function of the number
of extracted region proposals per image. The proposed method, using the selective 
search \cite{selectivesearchijcv2013} approach provides around 2000 regions per image as 
potential object locations. We rank the regions based on their scores. Jaccard IoU measure is employed to
neglect the regions with lower scores.

\begin{figure}
\centering
\includegraphics[width=0.51\textwidth]{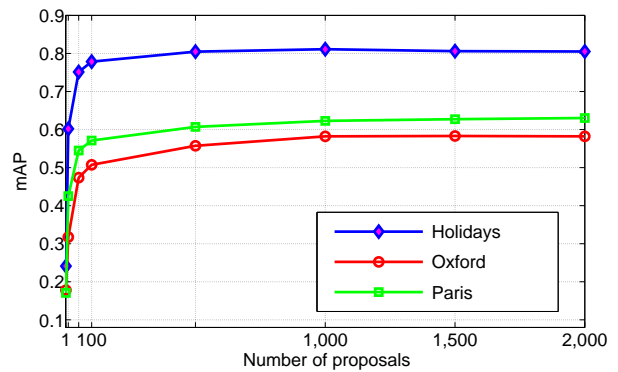} 
\caption{Retrieval performance on various databases considering different number of object proposals per image.}
\label{fig:mapvsN}
\end{figure}

The plot in Fig. \ref{fig:mapvsN} shows the mAP considering only the top N proposals for 
three datasets. It can be observed that with as minimum as $100$ proposals itself, the
proposed approach achieves very competitive results on all the databases. The approach can be
computationally very efficient by considering only these top scored CNN activations.

\subsection{Fine Tuning}
\begin{figure}[h]
\centering
\includegraphics[width=0.51\textwidth]{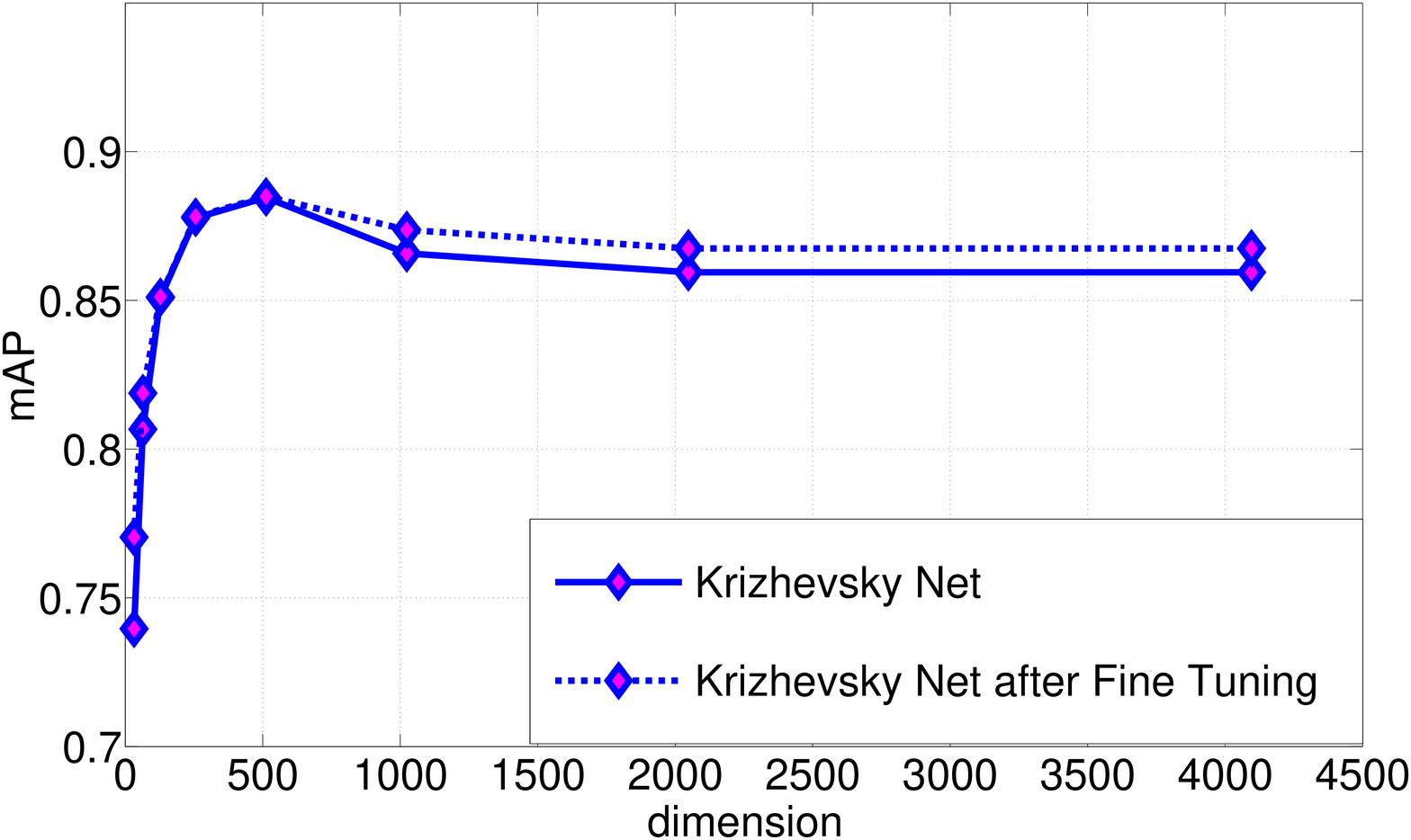} 
\caption{Retrieval performance on Holidays dataset with the object level deep features obtained after fine tuning the net.}
\label{fig:HD_Pretraining}
\end{figure}

\begin{figure}[h]
\centering
\includegraphics[width=0.51\textwidth]{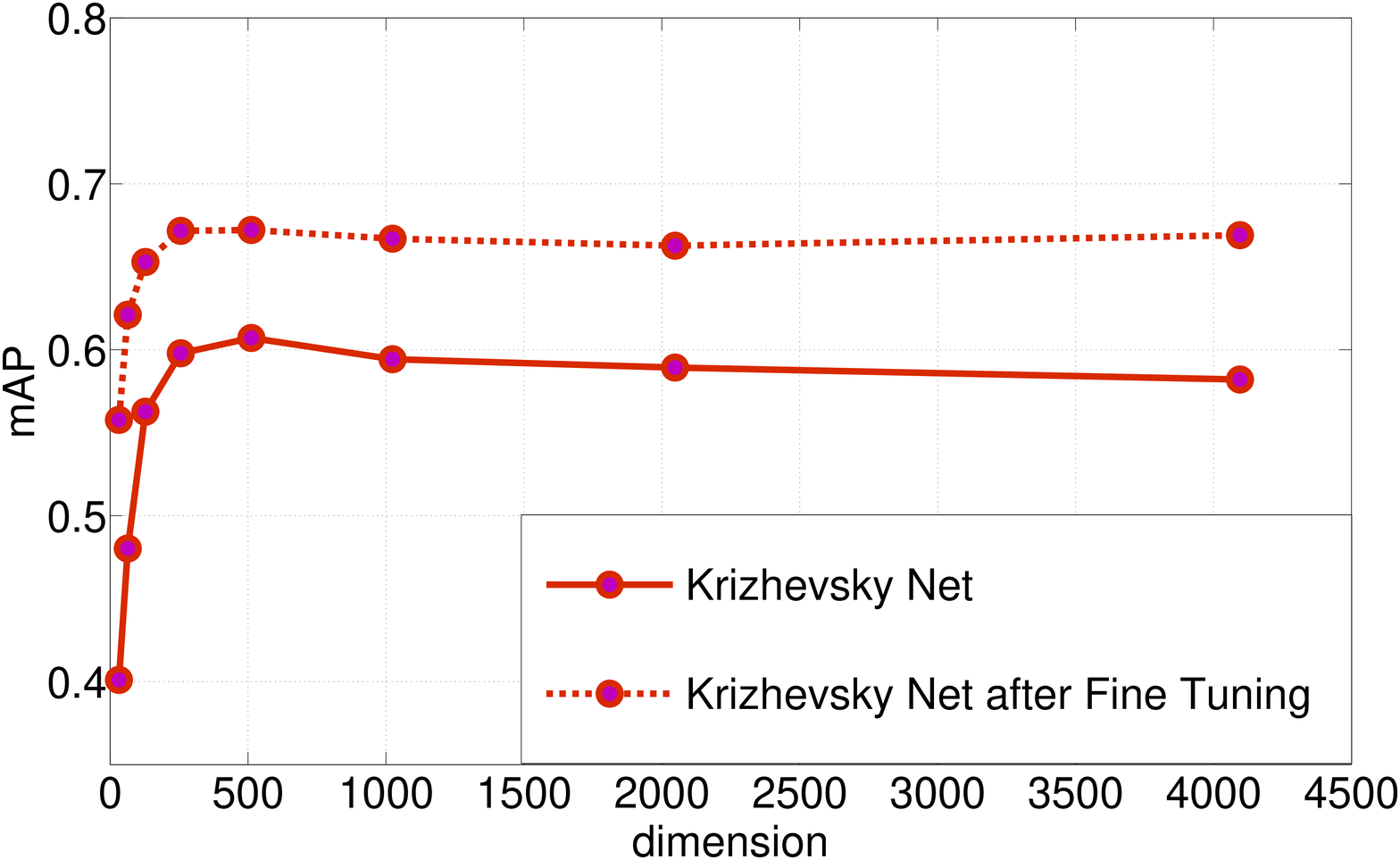}
\caption{Retrieval performance on Oxford5K dataset with the object level deep features obtained after fine tuning the net.}
\label{fig:OX_Pretraining}
\end{figure}

\begin{figure}
\centering
\includegraphics[width=0.51\textwidth]{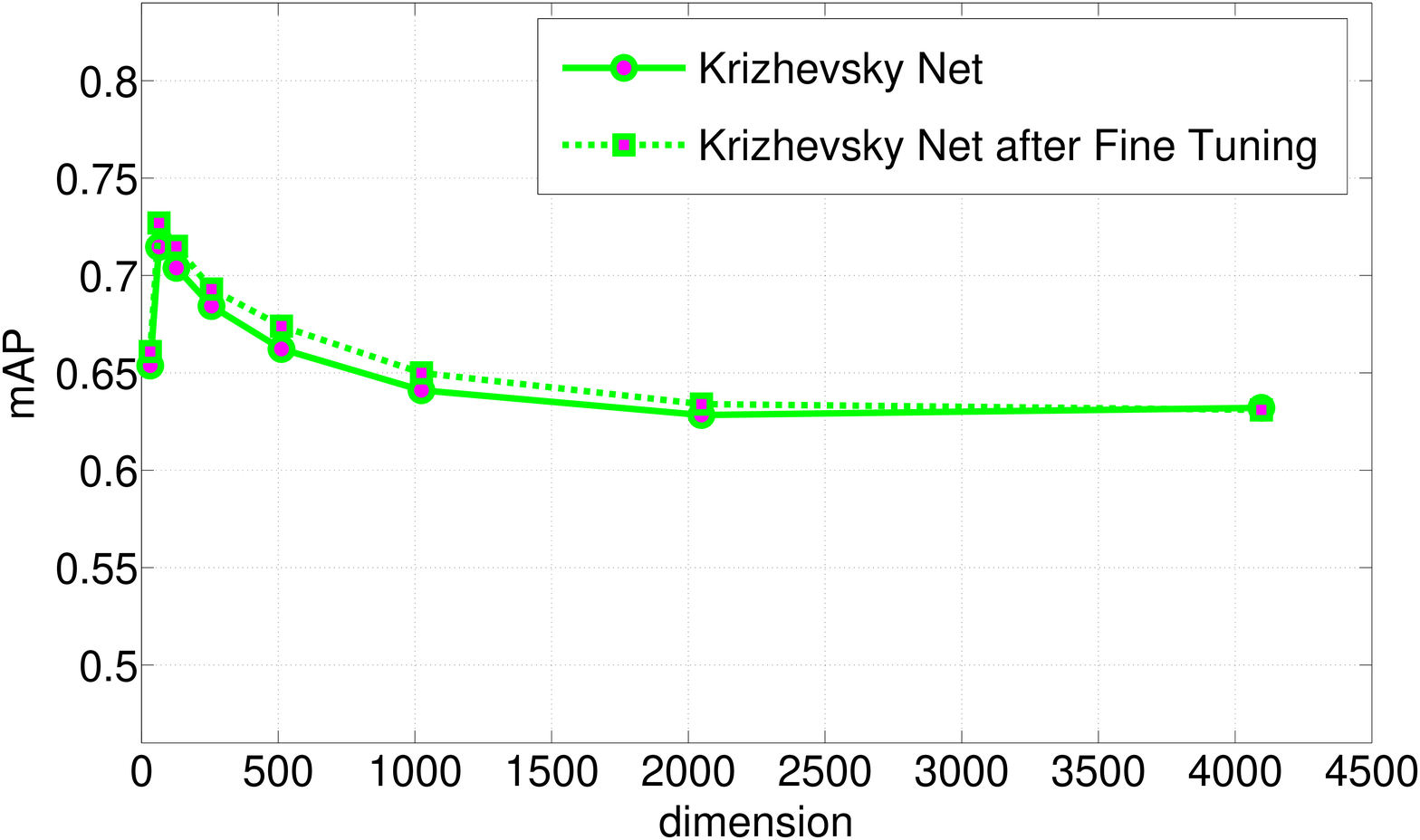} 
\caption{Retrieval performance on Paris6K dataset with the object level deep features obtained after fine tuning the net.}
\label{fig:PR_Pretraining}
\end{figure}

\begin{figure}[h]
\centering
\includegraphics[width=0.51\textwidth]{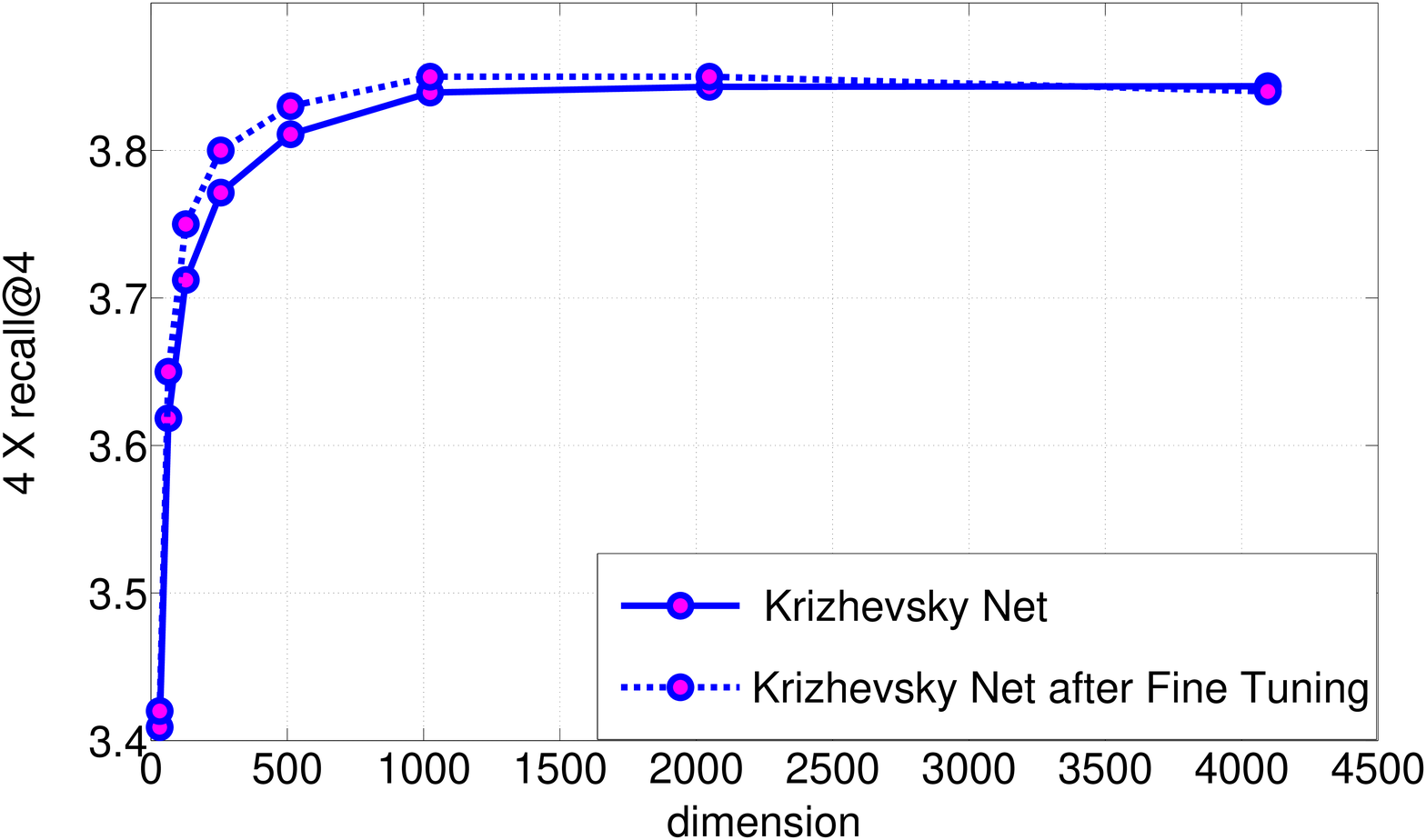}
\caption{Retrieval performance on UKB dataset with the object level deep features obtained after fine tuning the net.}
\label{fig:UKB_Pretraining}
\end{figure}

Apart from using the pre-trained model of the Alexnet CNN~\cite{deepcnnnips2012}, we have also fine tuned it with a very small
amount of training data from the dataset under investigation. For each dataset, 
we have presented around $100,000$ patches populated equally across all the classes, as the training data for fine
tuning. The final fully connected layer ($fc8$) is adjusted to have as many units as the number of classes in the corresponding dataset.
The learning rate is kept very low for the initial layers compared to $fc8$. The results for these experiments are presented in  
Figs.~\ref{fig:HD_Pretraining}, \ref{fig:OX_Pretraining}, \ref{fig:PR_Pretraining} and \ref{fig:UKB_Pretraining}.
Fine tuning of the model is observed to improve the
representation for most of the databases.


\newpage
\section{Conclusion}
\label{sec:conclu}

In this paper, we have demonstrated the effectiveness of the objectness prior on the Convolutional Neural Network (CNN)
activations of image regions, to impart invariance to the common geometric transformations such as
translation, scaling and rotation. The object patches are extracted at multiple scales in an efficient non-exhaustive manner.
The activations obtained for the extracted patches are aggregated in an order less manner to obtain an invariant image representation. 
Without incorporating any spatial information, our representation exhibits the state of the art performance for the 
image retrieval application with compact codes of dimensions less than 4096.
The binary representation with a memory footprint as small as $2Kbits$ per image also performs
on par with the state of the art. Our proposed method achieves these results with only $100-500$
region proposals per image, making it computationally non-intensive. The database specific
fine tuning of the learned model is observed to improve the representation
using minimal training data.
\section{Acknowledgements}
We gratefully acknowledge the support of NVIDIA Corporation with the donation of the K40 GPU used for this research.

\newpage

{\small
\bibliographystyle{ieee}
\bibliography{oprcnn}
}

\end{document}